\documentclass{article} 
\usepackage{iclr2017_conference,times}
\usepackage{hyperref}
\usepackage{url}
\usepackage{amssymb}
\usepackage{amsmath}
\usepackage[]{algorithm2e}
\usepackage{graphicx}
\usepackage{breqn}
\usepackage{sutton_math_symbols}
\usepackage{floatrow}
\usepackage{subcaption}
\newfloatcommand{capbtabbox}{table}[][\FBwidth]

\usepackage{color}

\usepackage{enumitem}
\setenumerate{itemsep=4pt,topsep=0pt,parsep=0pt,partopsep=0pt,leftmargin=2em}

\newcommand{\Mfvi}{{Decoupled Mean-Field Variational Inference}\xspace}
\newcommand{\mfvi}{{DMFVI}\xspace}
\newcommand{\plda}{\textsc{ProdLDA}\xspace}
\newcommand{\NVI}{\textsc{AEVB}\xspace}
\newcommand{\avitm}{\textsc{AVITM}\xspace}

\title{Autoencoding Variational Inference \\ For Topic Models}

\author{Akash Srivastava \\
Informatics Forum, University of Edinburgh\\
10, Crichton St\\
Edinburgh, EH89AB, UK \\
\texttt{akash.srivastava@ed.ac.uk} \\
\And
Charles Sutton\thanks{
Additional affiliation: Alan Turing Institute, British Library, 96 Euston Road, London NW1 2DB} \\
Informatics Forum, University of Edinburgh\\
10, Crichton St\\
Edinburgh, EH89AB, UK \\
\texttt{csutton@inf.ed.ac.uk} 
}

\iclrfinalcopy 

\begin{document}

\maketitle

\begin{abstract}
Topic models are one of the most popular methods for learning representations
of text, but a major challenge is that  any change to the topic model
requires  mathematically deriving a new inference algorithm.
A promising approach to address this problem is 
autoencoding variational Bayes (AEVB),
but it has proven difficult to apply to topic models in practice. We present what is to our knowledge
the first effective
AEVB
based inference method for latent Dirichlet allocation (LDA), which we call Autoencoded Variational Inference For Topic Model (AVITM).
This model tackles the problems caused for \NVI by the Dirichlet prior and by component collapsing.
We find that \avitm matches traditional methods in accuracy with much better inference time.
Indeed, because of the inference network, we find that it is unnecessary to pay the computational
cost of running
variational optimization on test data.
Because \avitm is black box, it is readily applied to new topic models.
As a dramatic illustration of this, we present a new topic model called ProdLDA,
that replaces the mixture model in LDA with a product
of experts. By changing only one line of code from LDA, we find that ProdLDA
yields much more interpretable topics, 
even if LDA is trained via collapsed Gibbs sampling.
\end{abstract}

\section{Introduction}

Topic models \citep{blei12topic} are among the most widely used models for learning
unsupervised representations of text, with hundreds of different model variants
in the literature, and have have found applications ranging from the exploration of the scientific literature
\citep{blei07science} to computer vision \citep{fei2005bayesian}, bioinformatics \citep{rogers2005latent}, and archaeology \citep{mimno09pompeii}.
A  major challenge in applying topic models and developing new models is the
computational cost of computing the posterior distribution. 
Therefore a large body of work has considered approximate inference methods,
the most popular methods being variational methods, especially mean field methods,
and Markov chain Monte Carlo, particularly methods based on collapsed Gibbs sampling.

Both mean-field and collapsed Gibbs have the drawback that applying them to new topic models,
even if there is only a small change to the modeling assumptions, requires re-deriving
the inference methods, which can be mathematically arduous and time consuming,
and limits the ability of practitioners to freely explore the space of different modeling
assumptions.  This has motivated the development of black-box inference methods
\citep{ranganath2014black,mnih2014neural,kucukelbir2016automatic,kingma2013auto} which require
only very limited and easy to compute information from the model, and hence
can be applied automatically to new models given a simple declarative specification
of the generative process.

Autoencoding variational Bayes (\NVI) \citep{kingma2013auto,rezende2014stochastic} is a particularly natural choice for topic models,
because it trains an \emph{inference network} \citep{dayan1995helmholtz}, a neural network that directly maps 
a document to an approximate posterior distribution, without the need to run further
variational updates. This is intuitively appealing 
because in topic models, we expect the mapping from documents to posterior distributions to be well behaved,
that is, that a small change in the document will produce only a small change in topics.
This is exactly the type of mapping that a universal function approximator like a neural
network should be good at representing. 
Essentially, the inference network learns to mimic the effect of probabilistic inference,
so that on test data, we can enjoy the benefits of probabilistic modeling without  
paying a further cost for inference.

However, despite some notable successes for latent Gaussian models, black box inference
methods are significantly more challenging to apply to topic models.
For example, in initial experiments, we tried to apply ADVI \citep{kucukelbir2016automatic}, a recent black-box
variational method, but it was difficult to obtain any meaningful topics.
Two main challenges are: first, the Dirichlet
prior is not a location scale family, which hinders reparameterisation, and second,
the well known problem of component collapsing \citep{dinh2016training}, in which the inference network becomes
stuck in a bad local optimum in which all topics are identical.

In this paper, we present what is, to our knowledge, the first effective \NVI inference method for topic models, which we call Autoencoded Variational Inference for Topic Models or \avitm \footnote{Code available at\\ \url{https://github.com/akashgit/autoencoding_vi_for_topic_models}}.   On several data sets,
we find that \avitm yields topics of equivalent quality to standard mean-field 
inference, with a large decrease in training time. We also find that the inference
network learns to mimic the process of approximate inference highly accurately,
so that it is not necessary to run variational optimization at all on test data.

But perhaps more important is that \avitm is a black-box method that is easy to apply
to new models.
To illustrate this, we present a new topic model, called ProdLDA, in which the distribution
over individual words is a product of experts rather than the mixture model used in LDA.
We find that ProdLDA consistently produces better topics than standard LDA, whether measured
by automatically determined topic coherence or qualitative examination.
Furthermore, because we perform probabilistic inference using a neural network,
we can fit a topic model on roughly a one million documents in under 80 minutes on a single GPU,
and because we are using a black box inference method, implementing ProdLDA requires a change
of \emph{only one line of code} from our implementation of standard LDA.

To summarize, the main advantages of our methods are:
\begin{enumerate}
  \item \emph{Topic coherence:} ProdLDA returns consistently better topics than LDA, even when
  LDA is trained using Gibbs sampling.
  \item \emph{Computational efficiency:} Training \avitm is fast and efficient like standard mean-field. On new data, \avitm is much faster than standard mean field, because it requires
  only one forward pass through a neural network.
  \item \emph{Black box:} \avitm does not require rigorous mathematical derivations to handle changes in the model, and can be easily applied to a wide range of topic  models.
\end{enumerate}
Overall, our results suggest that \avitm is ready to take its place
alongside mean field and collapsed Gibbs as one  of the workhorse inference methods for
topic models.

\section{Background}
\label{sec:background}

To fix notation,  we begin by describing  topic modelling
and \avitm.

\subsection{Latent Dirichlet Allocation}

We describe the most popular topic  model,
latent Dirichlet allocation (LDA).
In LDA, each document of the collection is represented as a mixture of topics,
where each topic $\beta_k$ is a probability distribution over the vocabulary.
We also use $\beta$ to denote the  matrix $\beta = (\beta_1 \ldots \beta_K)$.
The generative process is then as described in Algorithm~\ref{algo:lda}.
Under this generative model, the marginal likelihood of a document $\wB$ is
\begin{dmath}
\centering
\label{eq:lda_1}
p(\wB|\alpha,\beta) = \int_\theta \left( \prod_{n=1}^N \sum_{z_n = 1}^k p(w_n|z_n,\beta)p(z_n|\theta)\right)p(\theta|\alpha)d\theta.
\end{dmath}
  \begin{algorithm}[t]
 \For {each document $\wB$}{
  Draw topic distribution $\theta \sim  \mbox{Dirichlet} (\alpha)$\;
  \For{each word at position $n$}{
   Sample topic $z_{n} \sim \mbox{Multinomial}(1, \theta)$\;
   Sample  word $w_{n} \sim \mbox{Multinomial}(1, \beta_{z_n})$\;
   }
 }
 \caption{LDA as a generative model.}\label{algo:lda}\label{fig:lds}
\end{algorithm}

Posterior inference over the hidden variables $\theta$ and $z$ is intractable due to the coupling between the $\theta$ and $\beta$ under the multinomial assumption \citep{dickey1983multiple}. 

\subsection{Mean Field and \NVI}

A popular approximation for efficient inference in topic models is mean field variational inference, which
breaks the coupling between $\theta$ and $z$ by introducing free variational parameters $\gamma$ over $\theta$ and $\phi$ over $z$ and dropping the edges between them. This results in an approximate variational posterior $q(\theta,z|\gamma,\phi)=q_\gamma(\theta) \prod_n q_\phi ( z_n)$, which is optimized to best approximate the true posterior $p(\theta,z|\wB,\alpha,\beta)$.
The optimization problem is to minimize
\begin{dmath}
\centering
\label{eq:opt1}
L(\gamma,\phi \mid \alpha,\beta)= D_{KL}\left[q(\theta,z|\gamma,\phi)|| p(\theta,z|\wB,\alpha,\beta)\right]-\log p(\wB|\alpha,\beta).
\end{dmath}
In fact the above equation is a lower bound to the marginal log likelihood, sometimes called an \emph{evidence lower bound (ELBO)}, a fact which can be easily verified by  multiplying and dividing \eqref{eq:lda_1} by the variational posterior and then applying 
Jensen's inequality on its logarithm.
Note that the mean field method optimizes over an independent set of variational parameters for each document. To emphasize this,
we will refer to this standard method by the non-standard name of \emph{\Mfvi} (\mfvi).

For LDA, this optimization has closed form coordinate descent equations due to the conjugacy between the Dirichlet and multinomial distributions.
Although this is a computationally convenient aspect of \mfvi, it also limits its flexibility. 
Applying \mfvi to new models relies on the practitioner's ability to derive the closed form updates, 
which can be impractical and sometimes impossible. 

\NVI \citep{kingma2013auto,rezende2014stochastic} is one of several recent methods that aims at ``black box'' inference methods to sidestep this issue.
First, rewrite the ELBO as

\begin{dmath}
\label{eq:elboKL}
L(\gamma,\phi \mid \alpha,\beta)= -D_{KL}\left[q(\theta,z|\gamma,\phi)|| p(\theta,z | \alpha)\right] + \mathbb{E}_{q(\theta,z|\gamma,\phi)} [\log p(\wB|z,\theta,\alpha,\beta)]
\end{dmath}
This form is intuitive. The first term  attempts to match the variational posterior over latent variables to the prior on the latent variables, while the second term ensures that the variational posterior favors values of the latent variables that are good
at explaining the data. By analogy to autoencoders, this second term is
referred to as a \emph{reconstruction term}.


What makes this method ``Autoencoding,'' and in fact the main difference from \mfvi, is the parameterization of the variational distribution.
In \NVI, the variational parameters are computed by using a neural network called an \emph{inference network}
that takes the observed data as input. For example, if the model prior $p(\theta)$ were Gaussian,
we might define the inference network as a feedforward neural network $(\mu(\wB), \vB(\wB)) = f(\wB, \gamma),$ where $\mu(\wB)$ and $\vB(\wB)$ are both vectors of length $k$,
and $\gamma$ are the network's parameters. Then we might choose a Gaussian variational
distribution $q_\gamma(\theta) = N(\theta; \mu(\wB), \mbox{diag}(\vB(\wB))),$ 
where diag$(\cdots)$ produces a diagonal matrix from a column vector.
The variational parameters $\gamma$ can then be chosen by optimizing the ELBO \eqref{eq:elboKL}.
Note that we have now, unlike \mfvi, coupled the variational parameters for different documents because they are all computed from the same neural network.
To compute the expectations with respect to $q$ in \eqref{eq:elboKL}, \cite{kingma2013auto,rezende2014stochastic} use a Monte Carlo estimator
which they call the ``reparameterization trick'' (RT; appears also in \cite{williams1992simple}).
In the RT, we define a variate $U$ with a simple distribution that is independent of all variational parameters,
like a uniform or standard normal, and a reparameterization function $F$ such that $F(U, \gamma)$ has
distribution $q_\gamma$. This is always possible, as we could choose $F$ to be the inverse cumulative distribution
function of $q_\gamma$, although we will additionally want $F$ to be easy to compute and differentiable.
If we can determine a suitable $F$, then we can approximate \eqref{eq:elboKL} by taking Monte Carlo samples of $U$, 
and optimize $\gamma$ using stochastic gradient descent.

\section{Autoencoding Variational Bayes In Latent Dirichlet Allocation}

Although simple conceptually, applying \NVI to topic models raises several practical challenges.
The first is the need to determine a reparameterization function for $q(\theta)$ and $q(z_n)$
to use the RT. The $z_n$ are easily dealt with, but $\theta$ is more difficult; if we choose 
$q(\theta)$ to be Dirichlet, it is difficult to apply the RT, whereas if we choose $q$ to 
be Gaussian or logistic normal, then the KL divergence in \eqref{eq:elboKL} becomes more problematic.
The second issue is the well known problem of component collapsing \citep{dinh2016training},
which a type of bad local optimum that is particularly endemic to \NVI and similar methods.
We describe our solutions to each of those problems in the next few subsections.

\subsection{Collapsing $\zB$'s}

Dealing with discrete variables like $\zB$ using reparameterization can be problematic, but fortunately
in LDA the variable $\zB$ can be conveniently summed out.
By collapsing $\zB$ we are left with having to sample from $\theta$ only, reducing
\eqref{eq:lda_1} to
\begin{dmath}
\centering
\label{eq:lda_new}
p(\wB|\alpha,\beta) = \int_\theta \left( \prod_{n=1}^N  p(w_n|\beta, \theta)\right)p(\theta|\alpha)d\theta.
\end{dmath}
where the distribution of $w_n|\beta, \theta$ is Multinomial(1, $\beta \theta$), 
recalling that $\beta$ denotes the matrix of all topic-word probability vectors.

\subsection{Working with Dirichlet Beliefs: Laplace Approximation}

LDA gets its name from the Dirichlet prior  on the topic proportions $\theta$,
and the choice of Dirichlet prior is important to obtaining interpretable topics \citep{wallach09}. 
But it is difficult to handle the Dirichlet within \NVI because it is difficult to develop an effective
reparameterization function for the RT.
Fortunately, a RT does exist for the Gaussian distribution and has been shown to perform quite well in the context of variational autoencoder (VAE) \citep{kingma2013auto}. 

We resolve this issue by constructing a Laplace approximation to the Dirichlet prior. Following \cite{mackay1998choice}, we do so in the softmax basis instead of the simplex. There are two benefits of this choice. First, Dirichlet distributions are unimodal in the softmax basis with their modes coinciding with the means of the transformed densities. Second, the softmax basis also allows for carrying out unconstrained optimization of the cost function without the simplex constraints. The
Dirichlet probability density function in this basis over the softmax variable $\hB$ is given by
\begin{dmath}
\centering
\label{eq:dir}
P( \theta(\hB)| \alpha)=\frac{\Gamma(\sum_k \alpha_k)}{\prod_k \Gamma(\alpha_k)}\prod_k \theta_{k}^{\alpha_k}g(\textbf{1$^T$} \hB).
\end{dmath}
Here $\theta = \sigma(\hB)$, where $\sigma(.)$ represents the softmax function. Recall that the Jacobian of $\sigma$ is proportional to $\prod_k\theta_k$ and $g(\cdot)$ is an arbitrary density that ensures integrability by constraining the redundant degree of freedom. We use the Laplace approximation of \citet{hennig2012kernel}, which has the property that the covariance matrix becomes diagonal for large $k$ (number of topics). This approximation to the Dirichlet prior $p(\theta|\alpha)$ is results in the 
distribution over the softmax variables $\hB$ as a multivariate normal
with mean $\mu_1$ and covariance matrix $\Sigma_1$ where
\begin{align}
\label{eq:dir_mu}
\mu_{1k} &= \log \alpha_k - \frac{1}{K}\sum_i \log \alpha_i \nonumber \\
\Sigma_{1kk} &= \frac{1}{\alpha_k}\left(1 - \frac{2}{K}\right) + \frac{1}{K^2}\sum_i\frac{1}{\alpha_k}.
\end{align}

Finally, we approximate $p( \theta| \alpha)$ in the simplex basis with $\hat{p}(\theta|\mu_1,\Sigma_1)=\mathcal{LN}(\theta|\mu_1,\Sigma_1)$ where $\mathcal{LN}$ is a logistic normal distribution with parameters ${\mu_1,\Sigma_1}$. Although we approximate the Dirichlet prior in LDA with a logistic normal, this  is \emph{not}
the same idea as a correlated topic model \citep{correlated-topic-model}, because we use a diagonal covariance matrix. Rather, it is an approximation
to standard LDA.

\subsection{Variational Objective}

Now we can write the modified variational objective function. We use a logistic normal variational distribution over $\theta$
with diagonal covariance.
More precisely, we define two inference networks as feed forward neural networks $f_\mu$ and $ f_\Sigma$ with parameters $\pmb \delta$;
the output of each network is a vector in $\R^K.$
Then for a document $\wB$, we define $q(\theta)$ to be logistic normal with mean $\mu_0 = f_\mu(\wB,\pmb \delta)$
and diagonal covariance $\pmb \Sigma_0= \mbox{diag}(f_\Sigma(\wB,\pmb \delta))$, where diag converts a column vector
to a diagonal matrix. Note that we can generate samples from $q(\theta)$
by sampling $\pmb \epsilon \sim \mathcal{N}(0,I)$ and computing $\theta = \sigma(\pmb \mu_0+\pmb \Sigma_0^{1/2} \pmb \epsilon)$.

We can now write the ELBO as
\begin{align}
\label{eq:newcost}
L(\pmb \Theta)= \sum_{d=1}^D \Bigg [-\bigg(\frac{1}{2}\Big \{tr(\pmb \Sigma_1^{-1}\pmb \Sigma_0) + (\pmb \mu_1-\pmb \mu_0)^T\pmb \Sigma_1^{-1}(\pmb \mu_1- \pmb\mu_0) -K + \log \frac{|\pmb \Sigma_1|}{|\pmb \Sigma_0|}\Big \}\bigg) \\
+ \mathbb{E}_{\pmb \epsilon \sim \mathcal{N}(0,I)} \bigg [\wB_d^\top \log  \Big ( \sigma(\pmb \beta) \sigma(\pmb \mu_0+\pmb \Sigma_0^{1/2} \pmb \epsilon) \Big )\bigg ] \Bigg], \nonumber
\end{align}

where $\pmb \Theta$ represents the set of all the model and variational parameters
and $\wB_1 \ldots \wB_D$ are the documents in the corpus. The first line in this equation arises
from the KL divergence between the two logistic normal distributions $q$ and $\hat{p}$, while the second line is the reconstruction error.

In order to impose the simplex constraint on the $\beta$ matrix during the optimization, we apply the softmax transformation. 
That is, each topic $\beta_{k} \in \R^V$ is unconstrained, and the notation $\sigma(\pmb \beta)$ means to apply the
softmax function separately to each column of the matrix $\beta$. Note that the mixture of multinomials for each word $w_n$ can then be written as
$
p(w_n|\beta, \theta) = \big[\sigma(\beta)\theta\big]_{w_n},
$
which explains the dot product in \eqref{eq:newcost}.
To optimize \eqref{eq:newcost}, we use stochastic gradient descent using Monte Carlo samples from $\pmb \epsilon$, following
the Law of the Unconscious Statistician.

\subsection{Training and Practical Considerations: Dealing with Component Collapsing}

\NVI is prone to component collapsing \citep{dinh2016training}, which is a particular type of local optimum 
very close to the prior belief, early on in the training.
As the latent dimensionality of the model is increased, the KL regularization in the variational objective dominates,
so that  the outgoing decoder weights collapse for the components of the latent variable that reach close to the prior and do not show any posterior divergence. 
In our case, the collapsing specifically occurs because of the inclusion of the softmax transformation to produce $\theta$.
The result is that the $k$ inferred topics are identical as shown in table \ref{tab:cc}.

We were able to resolve this issue by tweaking the optimization.
Specifically,  we train the network with the ADAM optimizer \citep{kingma2014adam} using high moment weight ($\beta 1$) and learning rate ($\eta$). Through training at higher rates, early peaks in the functional space can be easily avoided. The problem is that momentum based training coupled with higher learning rate causes the optimizer to diverge. While explicit gradient clipping helps to a certain extent, we found that batch normalization \citep{ioffe2015batch} does even better by smoothing out the functional space and hence curbing sudden divergence. 

Finally, we also found an increase in performance with dropout units when applied to $\theta$ to force the network to use more of its capacity. 

While more prominent in the \NVI framework, the collapsing can also occurs in \mfvi if the learning offset (referred to as the $\tau$ parameter \citep{hofmann1999probabilistic}) is not set properly. Interestingly, a similar learning offset or annealing based approach can also be used to down-weight the KL term in early iterations of the training to avoid local optima.

\section{\plda: Latent Dirichlet Allocation with Products of Experts}
In LDA, the distribution $p(\wB | \theta, \beta)$ is a mixture of multinomials. A problem with this assumption is that it can never make any predictions that are sharper than the components that are being mixed \citep{hinton2009replicated}. This can result in some topics appearing
that are poor quality and do not correspond well with human judgment. 
One way to resolve this issue is to replace this word-level mixture with a weighted product of experts which by definition is capable of making sharper predictions than any of the constituent experts \citep{hinton2002training}. In this section we present a novel topic model \plda that replaces the mixture assumption at the word-level in LDA with a weighted product of experts, resulting in a drastic improvement in topic coherence. 
This is a good illustration of the benefits of a black box inference method, like \avitm, to allow exploration of new models.

\subsection{Model}

The \plda model can be simply described as latent Dirichlet allocation where the 
word-level mixture over topics is carried out in natural parameter space, i.e. the topic matrix is not constrained to exist 
in a multinomial simplex prior to mixing.
In other words, the only changes from LDA are that $\beta$ is unnormalized, and that the conditional distribution of
$w_n$ is defined as $w_n | \beta, \theta \sim \text{Multinomial}(1, \sigma(\beta \theta))$.

The connection to a product of experts is straightforward, as for the multinomial, a mixture of natural parameters corresponds
to a weighted geometric average of the mean parameters. 
That is, consider two $N$ dimensional multinomials parametrized by mean vectors $\pB$ and $\qB$. 
Define the corresponding natural parameters as $\pB = \sigma(\pmb r)$ and $\qB = \sigma(\pmb s),$ and let $\delta \in [0, 1].$
It is then easy to show that
$$P\Big(\xB|\delta \pmb r + (1-\delta) \pmb s\Big) \propto \prod_{i=1}^N \sigma(\delta r_i + (1-\delta) s_i)^{x_i}
\propto \prod_{i=1}^N [r_i^\delta \cdot s_i^{(1-\delta)} ]^{x_i}.$$
So the \plda model can be simply described as a product of experts, that is, $p(w_n | \theta, \beta) \propto \prod_k p(w_n | z_n = k, \beta)^{\theta_k}$. \plda is an instance of the exponential-family PCA \citep{collins2001generalization} class, and relates to the exponential-family harmoniums \citep{welling2004exponential} but with non-Gaussian priors.

\section{Related Work}
\label{sec:related}
For an overview of topic modeling, see \citet{blei12topic}.
There are several examples of  topic models
based on neural networks and neural variational inference  \citep{hinton2009replicated,larochelle2012neural,mnih2014neural,miao2015neural} 
but we are unaware of methods that apply \NVI generically to a topic model specified by an analyst,
or even of a successful application of \NVI to the most widely used topic model, latent Dirichlet allocation.

Recently, \cite{miao2015neural} introduced a closely related model called the Neural Variational Document Model (NVDM).
This method uses a latent Gaussian distribution over topics, like probabilistic latent semantic indexing, and averages over topic-word distributions 
in the logit space. However, they do not use either of the two key aspects of our work: explicitly approximating the Dirichlet
prior using a Gaussian, or high-momentum training. In the experiments we show that these aspects lead to much improved training
and much better topics. 

\section{Experiments and Results}
\label{sec:exp}
Qualitative evaluation of topic models is a challenging task and consequently a large body of work has 
developed automatic evaluation metrics that attempt to match human judgment of topic quality. Traditionally, {perplexity} has been used to measure the goodness-of-fit of the model but it has been repeatedly shown that perplexity is not a good metric for qualitative evaluation of topics \citep{newman2010automatic}. Several new metrics of topic coherence evaluation have thus been proposed; see \citet{lau2014machine} for a comparative review. \citet{lau2014machine} showed that among all the competing metrics, normalized pointwise mutual information (NPMI) between all the pairs of words in a set of topics matches human judgment most closely, so we adopt it in this work.
We also report  perplexity, primarily as a way of  evaluating  the capability of different optimizers. Following standard practice \citep{blei2003latent}, for variational methods we use the ELBO to calculate perplexity. 
For \NVI methods, we calculate the ELBO using the same Monte Carlo approximation as for training.

We run experiments on both the \textit{20 Newsgroups} (11,000 training instances with 2000 word vocabulary) and \textit{RCV1 Volume 2} (~800K training instances with 10000 word vocabulary) datasets. Our preprocessing 
involves tokenization, removal of some non UTF-8 characters for 20 Newsgroups
and English stop word removal. 
We first compare our \avitm inference method
with the standard online mean-field variational inference \citep{hoffman2010online} and collapsed Gibbs sampling \citep{griffiths2004finding} on the LDA model. We use standard implementations of both methods, \texttt{scikit-learn} for DMFVI and \texttt{mallet} \citep{mccallum:mallet} for collapsed Gibbs.
Then we compare two autoencoding inference methods on three different topic models: standard LDA, \plda using our inference method and the Neural Variational Document Model (NVDM) \citep{miao2015neural}, using the inference described in the paper.\footnote{We have used both \url{https://github.com/carpedm20/variational-text-tensorflow} and the NVDM author's \citep{miao2015neural} implementation.}

\begin{table}[H]
\centering
\caption{Average topic coherence on the 20 Newsgroups dataset. Higher is better.}
\label{tab:results_20}
\begin{tabular}{|c|c|c|c|c|c|}
\hline
\textbf{\# topics} & \textbf{\begin{tabular}[c]{@{}c@{}}ProdLDA\\ VAE \end{tabular}} & \textbf{\begin{tabular}[c]{@{}c@{}}LDA\\ VAE \end{tabular}} & \textbf{\begin{tabular}[c]{@{}c@{}}LDA\\ DMFVI\end{tabular}} & \textbf{\begin{tabular}[c]{@{}c@{}}LDA\\ Collapsed Gibbs\end{tabular}} & \textbf{NVDM} \\ \hline
\textbf{50}        & \textbf{0.24}   & 0.11                                                      & 0.11                                                       & 0.17                                                                 & 0.08        \\ \hline
\textbf{200}       & \textbf{0.19}  & 0.11                                                       & 0.06                                                       & 0.14                                                                 & 0.06        \\ \hline
\end{tabular}
\end{table}
Tables \ref{tab:results_20} and \ref{tab:results_rcv1} show the average topic coherence values for all the models for two different settings of $k$, the number of topics. Comparing the different inference methods for LDA, we find that, consistent with previous work, collapsed Gibbs sampling
yields better topics than mean-field methods. Among the variational methods, we find that VAE-LDA model (\avitm) \footnote{We recently found that 'whitening' the topic matrix significantly improves the topic coherence for VAE-LDA. Manuscript in preparation.} yields similar
topic coherence and perplexity to the standard DMFVI (although in some cases, VAE-LDA yields significantly better topics).
However, \avitm is significantly faster to train than DMFVI. It takes 46 seconds on 20 Newsgroup compared to 18 minutes for DMFVI. Whereas for a million document corpus of RCV1 it only under 1.5 hours while \textit{scikit-learn's} implementation of DMFVI failed to return any results even after running for 24 hours.\footnote{Therefore, we were not able to report topic coherence for DMFVI on RCV1}

Comparing the new topic models than LDA, it is clear that \plda finds significantly better topics than LDA, even when trained by collapsed Gibbs sampling. 
To verify this qualitatively, we display examples of topics from all the models in Table~\ref{tab:rand}.
The topics from  ProdLDA appear visually more coherent than NVDM or LDA.
Unfortunately, NVDM does not perform comparatively to LDA for any value of $k$. To avoid any training dissimilarities we train all the competing models until we reach the perplexities that were reported in previous work.
These are reported in Table \ref{tab:perp}\footnote{We note that much recent work follows \citet{hinton2009replicated} in reporting perplexity
for the LDA Gibbs sampler on only a small subset of the test data. Our results are different because we use the entire test dataset.}.

\begin{table}[t]
\centering
\caption{Average topic coherence on the RCV1 dataset. Higher is better. Results not reported for LDA DMFVI, as inference failed to converge in 24 hours.}
\label{tab:results_rcv1}
\begin{tabular}{|c|c|c|c|c|c|}
\hline
\textbf{\# topics} & \textbf{\begin{tabular}[c]{@{}c@{}}ProdLDA\\ VAE \end{tabular}} & \textbf{\begin{tabular}[c]{@{}c@{}}LDA\\ VAE\end{tabular}} & \textbf{\begin{tabular}[c]{@{}c@{}}LDA\\ DMFVI\end{tabular}} & \textbf{\begin{tabular}[c]{@{}c@{}}LDA\\ Collapsed Gibbs\end{tabular}} & \textbf{NVDM} \\ \hline
\textbf{50}        & \textbf{0.14}    & 0.07                                                         & -                                                       & 0.04                                                                   & 0.07          \\ \hline
\textbf{200}       & \textbf{0.12}    & 0.05                                                         & -                                                       & 0.06                                                                   & 0.05          \\ \hline
\end{tabular}
\end{table}
 
\begin{table}[t]
\centering
\caption{Perplexity scores for 20 Newsgroups. Lower is better.}
\label{tab:perp}
\begin{tabular}{|c|c|c|c|c|c|}
\hline
\textbf{\# topics} & \textbf{\begin{tabular}[c]{@{}c@{}}ProdLDA\\ VAE \end{tabular}} & \textbf{\begin{tabular}[c]{@{}c@{}}LDA\\ VAE\end{tabular}} & \textbf{\begin{tabular}[c]{@{}c@{}}LDA\\ DMFVI\end{tabular}} & \textbf{\begin{tabular}[c]{@{}c@{}}LDA\\ Collapsed Gibbs\end{tabular}} & \textbf{NVDM} \\ \hline
\textbf{50}        & 1172             & 1059                                                         & 1046                                                         & \textbf{728}                                                           & 837           \\ \hline
\textbf{200}       & 1168             & 1128                                                         & 1195                                                         & \textbf{688}                                                           & 884           \\ \hline
\end{tabular}
\end{table}

A major benefit of \avitm inference is that it 
does not require running variational optimization, which can be costly, for new data.
Rather, the inference network can be used to obtain topic proportions for new documents for new data points
without running any optimization. We evaluate whether this approximation is accurate, i.e. whether the neural
network effectively learns to mimic probabilistic inference. 
We verify this by training the  model on the training set, then on the test set, holding the topics ($\beta$ matrix) fixed, 
and comparing the test  perplexity if we obtain topic proportions by running the inference neural network directly, or
by the standard method of variational optimization of the inference network on the test set.
As shown in Table \ref{tab:better}, the perplexity remains practically un-changed. 
The  computational benefits of this are remarkable.
On both the datasets, computing perplexity using the neural network takes well under a minute, while running
the standard variational approximation takes $\sim 3$ minutes even on the smaller 20 Newsgroups data.
\begin{table}[t]
\centering
\caption{Evaluation of inference network of VAE-LDA on 20 Newsgroups test set.
``Inference network only'' shows the test perplexity when the inference network is trained on the training set,
but no variational optimization is performed on the test set. ``Inference Network + Optimization'' shows the standard
approach of optimizing the ELBO on the test set. The neural network effectively
learns to approximate probabilistic inference effectively.}

\label{tab:better}
\begin{tabular}{|c|c|c|}
\hline
\textbf{\# Topics} & \textbf{Inference Network Only} & \textbf{Inference Network + Optimization} \\[2pt] \hline
\textbf{50}                                                   & 1172                                                                         & 1162                                                                     \\ \hline
\textbf{200}                                                  & 1168                                                                         & 1151                                                                     \\ \hline
\end{tabular}
\end{table}
Finally, we investigate the reasons behind the improved topic coherence in \plda. First, Table \ref{tab:comp} explores the 
effects of each of our two main ideas separately. In this table, ``Dirichlet'' means that the prior  is the Laplace
approximation to Dirichlet($\alpha=0.02$), while ``Gaussian'' indicates that we use a standard Gaussian as prior. `High Learning Rate'' training means we use $\beta 1>0.8$ and $0.1>\eta>0.001$\footnote{$\beta 1$ is the weight on the average of the gradients from the previous time step and $\eta$ refers to the learning rate.} with batch normalization,
whereas ``Low Learning Rate'' means $\beta 1>0.8$ and $0.0009>\eta>0.00009$ without batch normalization.
(For both parameters,  the precise value was chosen by Bayesian optimization. We found that these values in the "\textit{with BN}" cases were close to the default settings in the Adam optimizer.)
We find that the high topic coherence that we achieve in this work is only possible if we use both tricks together. In fact the high learning rates with momentum is required to avoid local minima in the beginning of the training and batch-normalization is required to be able to train the network at these values without diverging. If trained at a lower momentum value or at a lower learning rate \plda shows component collapsing. 
Interestingly, if we choose a Gaussian prior, rather than the logistic normal approximation used in ProdLDA or NVLDA, the model is easier to train even with low learning rate without any momentum or batch normalization.

\begin{figure}[h]
  \includegraphics[width=.8\linewidth]{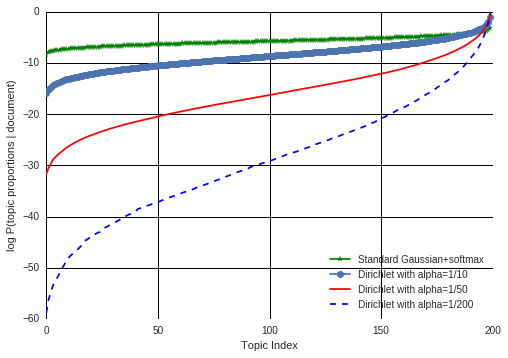}
  \caption{Effect of prior assumptions on $\theta$ on sparsity of $\theta$ in neural topic models.}
  \label{fig:sp}
\end{figure}

The main advantage of \avitm topic models as opposed to NVDM is that the Laplace approximation allows
us to match a specific Dirichlet prior of interest.
As pointed out by  \citet{wallach09},  the choice of Dirichlet hyperparameter is important to the topic quality of LDA.
Following this reasoning, we hypothesize that \avitm topics are higher quality than those of
NVDM because they are much more focused, i.e., apply to a  more specific subset  of documents
of interest. We provide  support for this hypothesis in 
Figure~\ref{fig:sp}, by evaluating the sparsity of the posterior proportions over topics,
that is, how many of the model's topics are typically used to explain each document.
In order to estimate the sparsity in topic proportions, we project samples from the Gaussian latent spaces of \plda and NVDM in the simplex and average them across documents. We compare the topic 
sparsity for the standard Gaussian prior  used by NVDM to the Laplace approximation
of Dirichlet priors with different hyperparameters.
Clearly the Laplace approximation to the Dirichlet prior significantly promotes sparsity,
providing support for our hypothesis that preserving the Dirichlet prior explains the
the increased topic coherence in our method.  
\begin{table}[h]
\centering
\caption{Average topic coherence for different choices of prior and optimization strategies of \plda on 20 Newsgroup for $k=50$.}
\label{tab:comp}
\resizebox{\textwidth}{!}{\begin{tabular}{|c|c|c|c|c|}
\hline
\textbf{}                & \textbf{\begin{tabular}[c]{@{}c@{}}Dirichlet\\ +High Learning Rate\end{tabular}} & \textbf{\begin{tabular}[c]{@{}c@{}}Dirichlet\\ +Low Learning Rate\end{tabular}} & \textbf{\begin{tabular}[c]{@{}c@{}}Gaussian Prior\\ +High Learning Rate\end{tabular}} & \textbf{\begin{tabular}[c]{@{}c@{}}Gaussian Prior\\ +Low Learning Rate\end{tabular}} \\ \hline
\textbf{Topic Coherence} & \textbf{0.24}                                                               & 0.016                                                                      & 0.08                                                                             & 0.08                                                                            \\ \hline
\end{tabular}}
\end{table}
\begin{table}[h]
\centering
\caption{Five \emph{randomly} selected topics from all the models.}
\label{tab:rand}
\resizebox{\textwidth}{!}{\begin{tabular}{|c|l|}
\hline
\textbf{Model}           & \multicolumn{1}{c|}{\textbf{Topics}}                                                                                                                                                                                                                                                                                                                                                                                       \\ \hline
\textbf{ProdLDA}         & \begin{tabular}[c]{@{}l@{}}motherboard meg printer quadra hd windows processor vga mhz connector\\ armenian genocide turks turkish muslim massacre turkey armenians armenia greek\\ voltage nec outlet circuit cable wiring wire panel motor install\\ season nhl team hockey playoff puck league flyers defensive player\\ israel israeli lebanese arab lebanon arabs civilian territory palestinian militia\end{tabular} \\ \hline
\parbox{0.75in}{\centering \textbf{LDA} \\ \textbf{NVLDA}}           & \begin{tabular}[c]{@{}l@{}}db file output program line entry write bit int return\\ drive disk get card scsi use hard ide controller one\\ game team play win year player get think good make\\ use law state health file gun public issue control firearm\\ people say one think life make know god man see\end{tabular}                                                                                                  \\ \hline
\parbox{0.75in}{\centering \textbf{LDA \\ DMFVI}}           & \begin{tabular}[c]{@{}l@{}}write article dod ride right go get night dealer like\\ gun law use drug crime government court criminal firearm control\\ lunar flyers hitter spacecraft power us existence god go mean\\ stephanopoulos encrypt spacecraft ripem rsa cipher saturn violate lunar crypto\\ file program available server version include software entry ftp use\end{tabular}                                   \\ \hline

\parbox{1in}{\centering\textbf{LDA \\ Collapsed Gibbs}} & \begin{tabular}[c]{@{}l@{}}get right back light side like see take time one\\ list mail send post anonymous internet file information user message\\ thanks please know anyone help look appreciate get need email\\ jesus church god law say christian one christ day come\\ bike dod ride dog motorcycle write article bmw helmet get\end{tabular}                                                                       \\ \hline
\textbf{NVDM}            & \begin{tabular}[c]{@{}l@{}}light die burn body life inside mother tear kill christian\\ insurance drug different sport friend bank owner vancouver buy prayer\\ input package interface output tape offer component channel level model\\ price quadra hockey slot san playoff jose deal market dealer\\ christian church gateway catholic christianity homosexual resurrection modem mouse sunday\end{tabular}            \\ \hline

\end{tabular}}
\end{table}

The inference network architecture can be found in figure \ref{fig:network} in the appendix.
\section{Discussion and Future Work}
\begin{table}[]
\centering
\caption{VAE-LDA fails to learn any meaningful topics when component collapsing occurs. The table shows five randomly sampled topics (, which are essentially slight variants of each other) from when the VAE-LDA model is trained without BN and high momentum training.}
\label{tab:cc}
\begin{tabular}{|l|}
\hline
\textbf{\begin{tabular}[c]{@{}l@{}}1. write article get thanks like anyone please know look one \\ 2. article write one please like anyone know make want get \\ 3. write article thanks anyone please like get one think look \\ 4. article write one get like know thanks anyone try need \\ 5. article write thanks please get like anyone one time make\end{tabular}} \\ \hline
\end{tabular}
\end{table}
We present what is to our knowledge the first effective \NVI inference algorithm for 
latent Dirichlet allocation.  Although this combination may seem simple in principle,
in practice  this method is difficult to train because of the Dirichlet prior
and because of the component collapsing problem.  By addressing both of these problems,
we presented a black-box inference method for topic models  with the notable advantage
that  the neural network allows computing topic proportions for new documents
without the need to run any variational optimization. As an illustration
of the advantages of black box inference techniques, we presented a new topic model, 
ProdLDA, which achieves significantly better topics than LDA, while requiring a change of only one line of code from \avitm for LDA.
Our results suggest that \avitm inference is ready to take its place
alongside mean field and collapsed Gibbs as one  of the workhorse inference methods for
topic models.
Future work could include extending our inference methods to handle 
dynamic and correlated topic models.

\subsubsection*{Acknowledgments}

We thank Andriy Mnih, Chris Dyer, Chris Russell, David Blei, Hannah Wallach, Max Welling, Mirella Lapata and Yishu Miao for helpful comments, discussions and feedback.





\bibliography{iclr2017_conference}

\begin{thebibliography}{33}
\providecommand{\natexlab}[1]{#1}
\providecommand{\url}[1]{\texttt{#1}}
\expandafter\ifx\csname urlstyle\endcsname\relax
  \providecommand{\doi}[1]{doi: #1}\else
  \providecommand{\doi}{doi: \begingroup \urlstyle{rm}\Url}\fi

\bibitem[Blei(2012)]{blei12topic}
David Blei.
\newblock Probabilistic topic models.
\newblock \emph{Communications of the ACM}, 55\penalty0 (4):\penalty0 77--84,
  2012.

\bibitem[Blei \& Lafferty(2006)Blei and Lafferty]{correlated-topic-model}
David~M. Blei and John~D. Lafferty.
\newblock Correlated topic models.
\newblock In \emph{Advances in Neural Information Processing Systems}, 2006.

\bibitem[Blei \& Lafferty(2007)Blei and Lafferty]{blei07science}
David~M. Blei and John~D. Lafferty.
\newblock A correlated topic model of science.
\newblock \emph{Annals of Applied Statistics}, 1\penalty0 (1):\penalty0 17--35,
  2007.

\bibitem[Blei et~al.(2003)Blei, Ng, and Jordan]{blei2003latent}
David~M Blei, Andrew~Y Ng, and Michael~I Jordan.
\newblock Latent dirichlet allocation.
\newblock \emph{Journal of machine Learning research}, 3\penalty0
  (Jan):\penalty0 993--1022, 2003.

\bibitem[Collins et~al.(2001)Collins, Dasgupta, and
  Schapire]{collins2001generalization}
Michael Collins, Sanjoy Dasgupta, and Robert~E Schapire.
\newblock A generalization of principal component analysis to the exponential
  family.
\newblock In \emph{Advances in Neural Information Processing Systems},
  volume~13, pp.\ ~23, 2001.

\bibitem[Dayan et~al.(1995)Dayan, Hinton, Neal, and Zemel]{dayan1995helmholtz}
Peter Dayan, Geoffrey~E Hinton, Radford~M Neal, and Richard~S Zemel.
\newblock The helmholtz machine.
\newblock \emph{Neural Computation}, 7\penalty0 (5):\penalty0 889--904, 1995.

\bibitem[Dickey(1983)]{dickey1983multiple}
James~M Dickey.
\newblock Multiple hypergeometric functions: Probabilistic interpretations and
  statistical uses.
\newblock \emph{Journal of the American Statistical Association}, 78\penalty0
  (383):\penalty0 628--637, 1983.

\bibitem[Dinh \& Dumoulin(2016)Dinh and Dumoulin]{dinh2016training}
Laurent Dinh and Vincent Dumoulin.
\newblock Training neural {Bayesian} nets.
\newblock
  \url{http://www.iro.umontreal.ca/~bengioy/cifar/NCAP2014-summerschool/slides/Laurent_dinh_cifar_presentation.pdf},
  August 2016.

\bibitem[Fei-Fei \& Perona(2005)Fei-Fei and Perona]{fei2005bayesian}
Li~Fei-Fei and Pietro Perona.
\newblock A {Bayesian} hierarchical model for learning natural scene
  categories.
\newblock In \emph{IEEE Computer Society Conference on Computer Vision and
  Pattern Recognition (CVPR'05)}, volume~2, pp.\  524--531. IEEE, 2005.

\bibitem[Griffiths \& Steyvers(2004)Griffiths and
  Steyvers]{griffiths2004finding}
Thomas~L Griffiths and Mark Steyvers.
\newblock Finding scientific topics.
\newblock \emph{Proceedings of the National academy of Sciences}, 101\penalty0
  (suppl 1):\penalty0 5228--5235, 2004.

\bibitem[Hennig et~al.(2012)Hennig, Stern, Herbrich, and
  Graepel]{hennig2012kernel}
Philipp Hennig, David~H Stern, Ralf Herbrich, and Thore Graepel.
\newblock Kernel topic models.
\newblock In \emph{AISTATS}, pp.\  511--519, 2012.

\bibitem[Hinton(2002)]{hinton2002training}
Geoffrey~E Hinton.
\newblock Training products of experts by minimizing contrastive divergence.
\newblock \emph{Neural computation}, 14\penalty0 (8):\penalty0 1771--1800,
  2002.

\bibitem[Hinton \& Salakhutdinov(2009)Hinton and
  Salakhutdinov]{hinton2009replicated}
Geoffrey~E Hinton and Ruslan~R Salakhutdinov.
\newblock Replicated softmax: an undirected topic model.
\newblock In \emph{Advances in Neural Information Processing Systems}, pp.\
  1607--1614, 2009.

\bibitem[Hoffman et~al.(2010)Hoffman, Bach, and Blei]{hoffman2010online}
Matthew Hoffman, Francis~R Bach, and David~M Blei.
\newblock Online learning for latent dirichlet allocation.
\newblock In \emph{Advances in Neural Information Processing Systems}, pp.\
  856--864, 2010.

\bibitem[Hofmann(1999)]{hofmann1999probabilistic}
Thomas Hofmann.
\newblock Probabilistic latent semantic indexing.
\newblock In \emph{Proceedings of the 22nd annual international ACM SIGIR
  conference on Research and development in information retrieval}, pp.\
  50--57. ACM, 1999.

\bibitem[Ioffe \& Szegedy(2015)Ioffe and Szegedy]{ioffe2015batch}
Sergey Ioffe and Christian Szegedy.
\newblock Batch normalization: Accelerating deep network training by reducing
  internal covariate shift.
\newblock pp.\  448--456, 2015.

\bibitem[Kingma \& Ba(2015)Kingma and Ba]{kingma2014adam}
Diederik Kingma and Jimmy Ba.
\newblock Adam: A method for stochastic optimization.
\newblock \emph{3rd International Conference on Learning Representations
  (ICLR)}, 2015.

\bibitem[Kingma \& Welling(2014)Kingma and Welling]{kingma2013auto}
Diederik~P Kingma and Max Welling.
\newblock Auto-encoding variational bayes.
\newblock \emph{The International Conference on Learning Representations
  (ICLR), Banff}, 2014.

\bibitem[Kucukelbir et~al.(2016)Kucukelbir, Tran, Ranganath, Gelman, and
  Blei]{kucukelbir2016automatic}
Alp Kucukelbir, Dustin Tran, Rajesh Ranganath, Andrew Gelman, and David~M Blei.
\newblock Automatic differentiation variational inference.
\newblock \emph{arXiv preprint arXiv:1603.00788}, 2016.

\bibitem[Larochelle \& Lauly(2012)Larochelle and Lauly]{larochelle2012neural}
Hugo Larochelle and Stanislas Lauly.
\newblock A neural autoregressive topic model.
\newblock In \emph{Advances in Neural Information Processing Systems}, pp.\
  2708--2716, 2012.

\bibitem[Lau et~al.(2014)Lau, Newman, and Baldwin]{lau2014machine}
Jey~Han Lau, David Newman, and Timothy Baldwin.
\newblock Machine reading tea leaves: Automatically evaluating topic coherence
  and topic model quality.
\newblock In \emph{EACL}, pp.\  530--539, 2014.

\bibitem[MacKay(1998)]{mackay1998choice}
David~JC MacKay.
\newblock Choice of basis for {Laplace} approximation.
\newblock \emph{Machine learning}, 33\penalty0 (1):\penalty0 77--86, 1998.

\bibitem[McCallum(2002)]{mccallum:mallet}
Andrew McCallum.
\newblock Mallet: A machine learning for language toolkit.
\newblock \url{http://mallet.cs.umass.edu}, 2002.

\bibitem[Miao et~al.(2016)Miao, Yu, and Blunsom]{miao2015neural}
Yishu Miao, Lei Yu, and Phil Blunsom.
\newblock Neural variational inference for text processing.
\newblock pp.\  1727--1736, 2016.

\bibitem[Mimno(2009)]{mimno09pompeii}
David Mimno.
\newblock Reconstructing {Pompeian} households.
\newblock In \emph{Applications of Topic Models Workshop, NIPS}, 2009.

\bibitem[Mnih \& Gregor(2014)Mnih and Gregor]{mnih2014neural}
Andriy Mnih and Karol Gregor.
\newblock Neural variational inference and learning in belief networks.
\newblock pp.\  1791--1799, 2014.

\bibitem[Newman et~al.(2010)Newman, Lau, Grieser, and
  Baldwin]{newman2010automatic}
David Newman, Jey~Han Lau, Karl Grieser, and Timothy Baldwin.
\newblock Automatic evaluation of topic coherence.
\newblock In \emph{Human Language Technologies: The 2010 Annual Conference of
  the North American Chapter of the Association for Computational Linguistics},
  pp.\  100--108. Association for Computational Linguistics, 2010.

\bibitem[Ranganath et~al.(2014)Ranganath, Gerrish, and
  Blei]{ranganath2014black}
Rajesh Ranganath, Sean Gerrish, and David~M Blei.
\newblock Black box variational inference.
\newblock In \emph{AISTATS}, pp.\  814--822, 2014.

\bibitem[Rezende et~al.(2014)Rezende, Mohamed, and
  Wierstra]{rezende2014stochastic}
Danilo~Jimenez Rezende, Shakir Mohamed, and Daan Wierstra.
\newblock Stochastic backpropagation and approximate inference in deep
  generative models.
\newblock pp.\  1278--1286, 2014.

\bibitem[Rogers et~al.(2005)Rogers, Girolami, Campbell, and
  Breitling]{rogers2005latent}
Simon Rogers, Mark Girolami, Colin Campbell, and Rainer Breitling.
\newblock The latent process decomposition of cdna microarray data sets.
\newblock \emph{IEEE/ACM Transactions on Computational Biology and
  Bioinformatics (TCBB)}, 2\penalty0 (2):\penalty0 143--156, 2005.

\bibitem[Wallach et~al.(2009)Wallach, Mimno, and McCallum]{wallach09}
Hanna Wallach, David Mimno, and Andrew McCallum.
\newblock Rethinking {LDA}: Why priors matter.
\newblock In \emph{NIPS}, 2009.

\bibitem[Welling et~al.(2004)Welling, Rosen-Zvi, and
  Hinton]{welling2004exponential}
Max Welling, Michal Rosen-Zvi, and Geoffrey~E Hinton.
\newblock Exponential family harmoniums with an application to information
  retrieval.
\newblock In \emph{Advances in Neural Information Processing Systems},
  volume~4, pp.\  1481--1488, 2004.

\bibitem[Williams(1992)]{williams1992simple}
Ronald~J Williams.
\newblock Simple statistical gradient-following algorithms for connectionist
  reinforcement learning.
\newblock \emph{Machine Learning}, 8\penalty0 (3-4):\penalty0 229--256, 1992.

\end{thebibliography}
\bibliographystyle{iclr2017_conference}

\appendix
\section{Network Architecture}
\begin{figure}[h]
  \includegraphics[width=.7\linewidth]{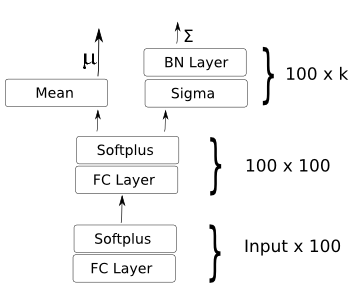}
  \caption{Architecture of the inference network used in the experiments.}
  \label{fig:network}
\end{figure}
\end{document}